\documentclass[final]{cvpr}

\usepackage{times}
\usepackage{epsfig}
\usepackage{graphicx}
\usepackage{amsmath}
\usepackage{amssymb}
\usepackage{epstopdf}
\usepackage{gensymb}

\usepackage[pagebackref=true,breaklinks=true,colorlinks,bookmarks=false]{hyperref}

\def\cvprPaperID{****} 
\def\confYear{CVPR 2021}

\begin{document}

\title{Evaluation of deep lift pose models for 3D rodent pose estimation based on geometrically triangulated data}

\author{Indrani Sarkar$^{1}$ \thanks{Equal contribution.} ~~~~ Indranil Maji$^1$\footnotemark[1] ~~~~ Charitha Omprakash$^2$\footnotemark[1] \\ 
Sebastian Stober$^1$ ~~~~ Sanja Mikulovic$^2$ ~~~~ Pavol Bauer$^{2}$\\
$^1$ Otto von Guericke University , Germany\\
$^2$ Leibniz Institute for Neurobiology, Germany\\
{\tt pavol.bauer@lin-magdeburg.de}}


\maketitle

\begin{abstract}
The assessment of laboratory animal behavior is of central interest in modern neuroscience research.
Behavior is typically studied in terms of pose changes, which are ideally captured in three dimensions. 
This requires triangulation over a multi-camera system which view the animal from different angles.
However, this is challenging in realistic laboratory setups due to occlusions and other technical constrains.
Here we propose the usage of lift-pose models that allow for robust 3D pose estimation of freely moving rodents from a single view camera view. 
To obtain high-quality training data for the pose-lifting, we first perform geometric calibration in a camera setup involving bottom as well as side views of the behaving animal.
We then evaluate the performance of two previously proposed model architectures under given inference perspectives and conclude that reliable 3D pose inference can be obtained using temporal convolutions.
With this work we would like to contribute to a more robust and diverse behavior tracking of freely moving rodents for a wide range of experiments and setups in the neuroscience community.
\end{abstract}

\section{Introduction}

A central goal of modern neuroscience research is to measure and quantify behavior of laboratory animals in order to enable correction studies to neuronal activity.

\begin{figure}[htb!] 
\centering
\includegraphics[width=0.95\linewidth]{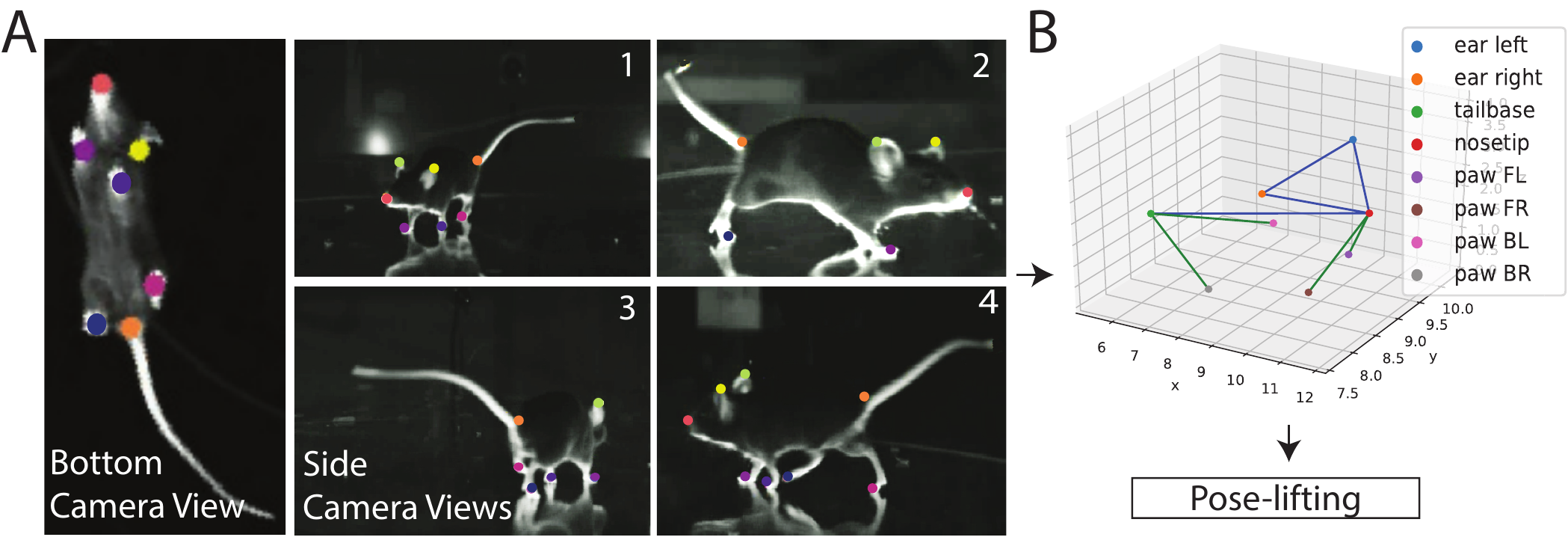}
\caption{A: Exemplary input data of unconstrained mouse behavior tracked with 2D keypoint detectors from 5 perspectives. B: Outline of the inference pipeline; first, we obtain high-quality triangulation data from the 5-camera setup. This data is then used to train the lift pose model for 3D inference from single camera views. }
\label{fig1}
\end{figure}


Animal pose estimation in two dimensions has been recently made possible using convolutional neural networks that allow keypoint detections throughout a recorded video based on the training data labeled by the user \cite{deeplabcut, pereira_fast_2019}.
If the goal is to obtain the animal pose in three dimensions, it is possible to combine the 2D keypoint detections from multiple synchronously operated cameras, and triangulate the points into a global 3D space \cite{hartley_multiple_2004}.
However, prior to the triangulation, the cameras need to be calibrated using a target, which is often an erroneous process in practice that can result in skewed projections of points into the global space.
Moreover, during many experiments in behavioral neuroscience, it is difficult to establish continuous 3D tracking of keypoints over time as often it can not be guaranteed that two or more cameras have a view on all tracked body parts.
This can occur due to different reasons; difficulties in mounting the cameras in the desired angles, occlusion of objects or others animals, and self occlusion of the animal itself.
To overcome such shortcomings for 3D pose estimation in humans, lift pose models have been developed, which aim to infer the 3D pose directly from a single camera view \cite{simplebaseline, videopose, cao2017realtime,sun_view-invariant_2020}.


This work contains two contributions. First, we present a simple and robust procedure for triangulation of freely moving rodents from multiple cameras that are orthogonal towards a camera positioned underneath the plane of movement (Figure \ref{fig2}).
Second, we evaluate two model architectures that have been proposed for 3D pose lifting of human poses previously, one a linear residual network \cite{simplebaseline} and the other a dilated temporal convolutional residual network \cite{videopose}, aiming to establish which of both models works better for our triangulated rodent pose data. 
Moreover, we evaluate the choice of the temporal window setting on the performance of the temporal convolutional model as well as the choice of the viewing perspective on the performance of both models.
The aim of this work is to pave the way for robust detection of 3D rodent poses from single poses, allowing for studies of behavior in complex laboratory as well as naturalistic environments.

\section{Related work}
Previously 3D pose estimation on humans was proposed by Martinez \etal \cite{simplebaseline} where a simple linear residual network was trained on HumanEva and Human3.6M dataset. This architecture has already been evaluated by Gosztolai \etal \cite{liftpose} for 3D pose estimation in freely moving monkeys as well as rodents in a constrained behavioral setup. Pose lifting of human data using temporal 1-dimensional dilated convolutional neural networks was previously proposed by Pavllo \etal \cite{videopose}. View-invariant human pose estimation from embedding spaces was recently proposed by Sun \etal \cite{sun_view-invariant_2020}.
Approaches to triangulation of animals include Anipose \cite{anipose} and more recently the DANNCE framework that employs 3D convolutions for improving detections of 3D poses captured from a multi-camera setup \cite{dunn_geometric_2021}.


\section{Geometric camera calibration}

\begin{figure}[ht!] 
\centering
\includegraphics[width=2.3in]{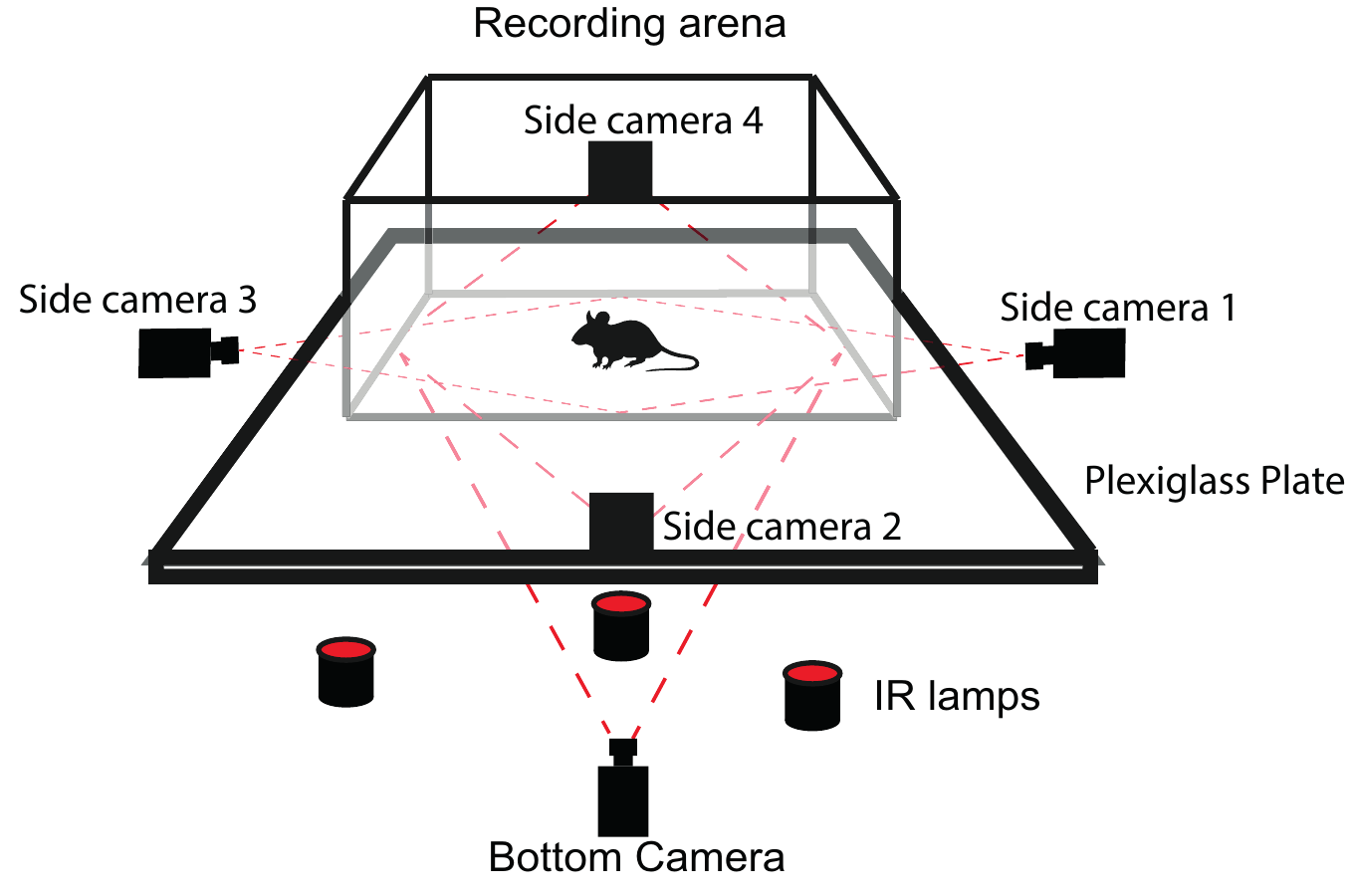}
\caption{Multi camera setup for geometric triangulation.}
\label{fig2}
\end{figure}

Classical triangulation of points is based on the projection of 2D planes onto a global 3D space. This requires calibration with a flat calibration target, typically a checker-board, that needs to be visible to all cameras simultaneously in order to identify the direction of the projection.
Here we propose a simple and robust triangulation process based on orthogonality of cameras that are unable to view the classical calibration target simultaneously in their spatial configuration (Figure \ref{fig2}).

This approach is motivated by the fact that the most variability of rodent movement is visible from the bottom perspective \cite{luxem_identifying_2020}, from which the $x,y$ position of many body parts can be directly detected given reliable 2d pose estimation.
The additional orthogonal cameras are then needed to determine the $z$ coordinate for each keypoint, which can be given by the evaluation of a polynomial function determined during the calibration process.

To perform the experiments, we let one C57BL/6J wildtype animal freely explore a circular arena with a transparent Plexiglas floor.
We recorded 60 minutes of unconstrained behavior using 5 synchronous cameras operating at 50 Hz that have been mounted as shown in Figure \ref{fig2}.
We then labeled 8 body parts in $1e3$ images and trained a ResNet CNN model provided in the \textit{DeepLabCut 2.2} \cite{deeplabcut} to detect the keypoints from the bottom view as well as side view recordings (Figure \ref{fig1}).
The image augmentation routines supplied in the \textit{DeepLabCut} package were used to improve the generalization of the network. 
The keypoint coordinates were egocentrically aligned, so that the nosetip and tailbase marker form a vector parallel to the $x$-axis  and the center of the animal body is approximately at the origin of the $x$ and $y$ axes.

\begin{figure}[ht!] 
\centering
\includegraphics[width=0.95\linewidth]{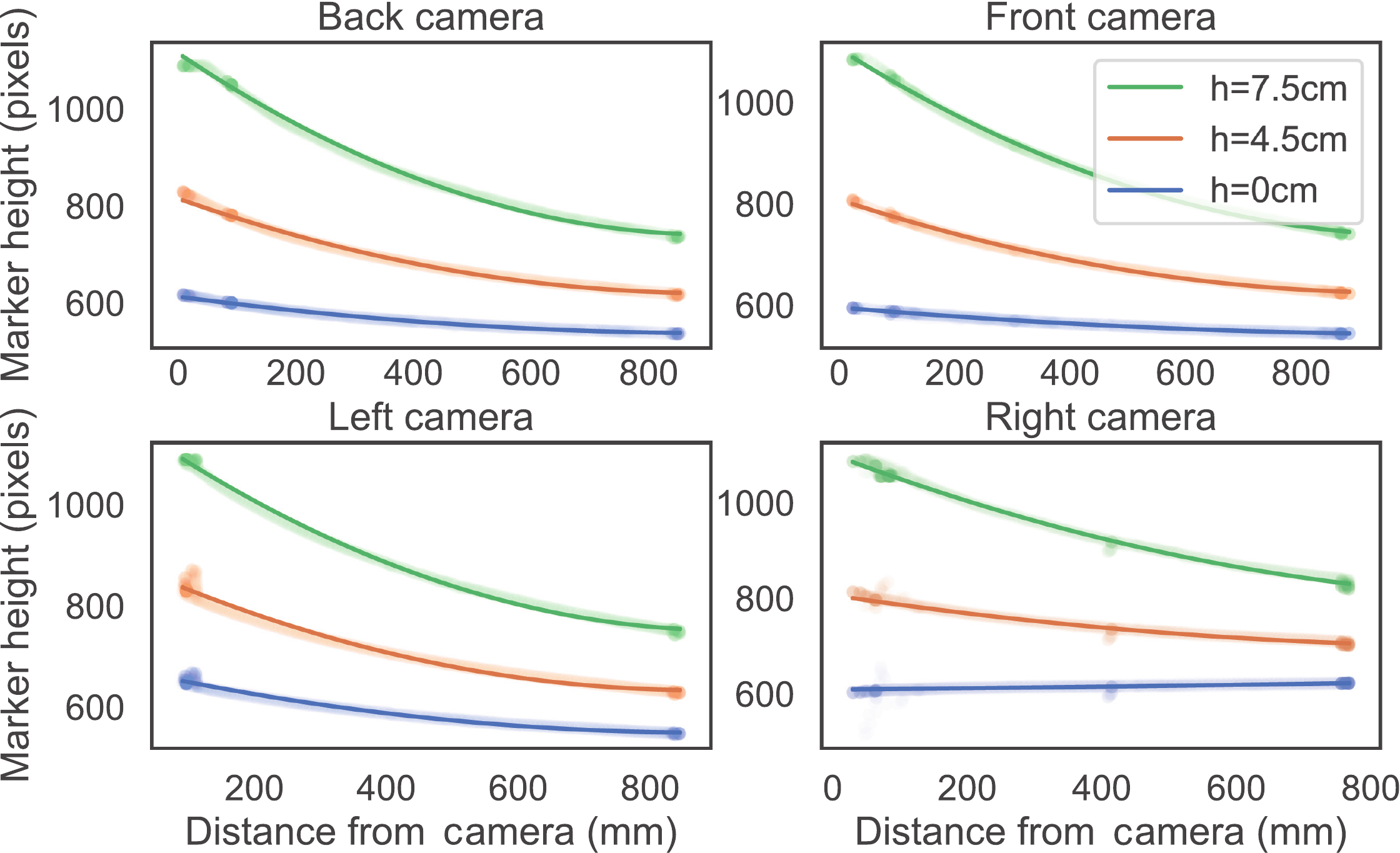}
\caption{Polynomial fitting function used for translating the $y$ coordinate of the calibration marker heights to the physical height in the arena, given the distance of the calibration target to the camera.}
\label{fig3}
\end{figure}

The triangulation procedure was performed as follows. The algorithm iterated over the tuples of detected $x,y$ coordinates obtained from the bottom perspective. For each timestep and keypoint, the algorithm then searched for detections from the side camera that were obtained from the ResNet at a high accuracy ($p > 0.95$). 
The height of the pixel on the side frame was converted into a physical $z$ value via the evaluation of a polynomial function, which was obtained a priori using a cylindrical calibration object with physical markers at three heights that was moved over the arena (Figure \ref{fig3}). 
If more then one side camera reliably detected a keypoint, an average of the detections was assigned to the $z$ value of the particular keypoint.
In total, in about $86\%$ of the time frames all body parts could be detected in one of the side cameras. The remaining points were interpolated using an exponentially weighted moving average filter. Finally, the mean centered and z-scored values have been used for training of the pose-lifting models.

\section{Lifting models}

\subsection{Linear Model }
\begin{figure}[ht!] 
\centering
\includegraphics[width=0.95\linewidth]{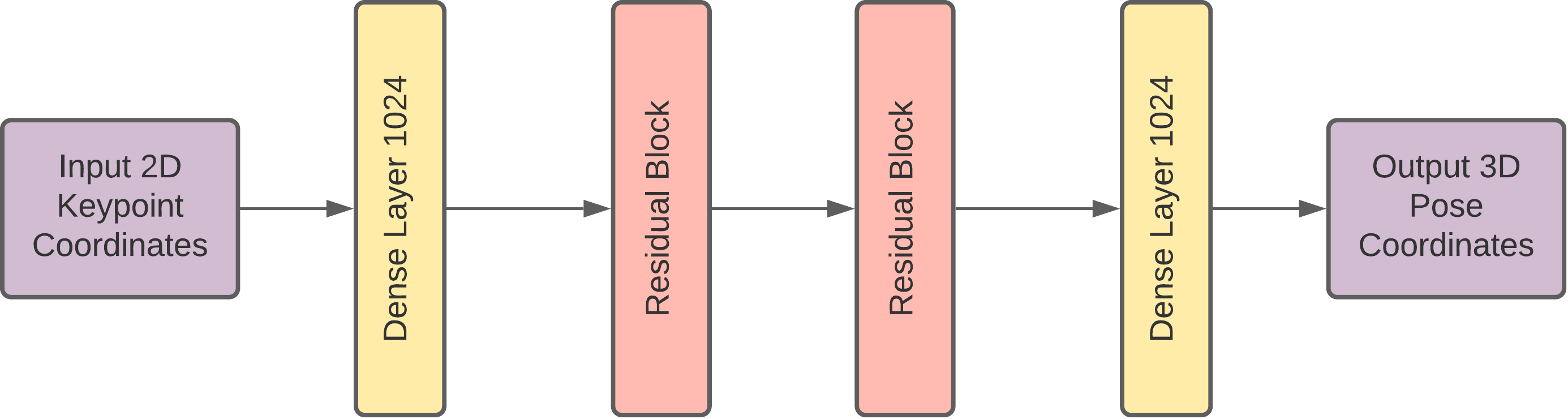}
\caption{Main Model Architecture of the \textit{Linear model}.}
\label{fig4}
\end{figure}

The \textit{Linear model} \cite{simplebaseline} architecture consists of a linear layer followed by a batchnorm layer, a ReLU activation function and dropout with 0.25 probability (Figure \ref{fig4}). The linear layer is then followed by a residual block consisting of two linear layers followed by batchnorm layer, ReLU and dropout, where the input and output of this block are connected by a residual connection. 

We used the MPJPE (Mean Per Joint Position Error) to compute the loss at every iteration,
\begin{equation}
 \text{MPJPE} = \frac{1}{N} \frac{1}{K} \sum_{i=1}^{N}\sum_{k=1}^{K} \parallel f(x) - y\parallel,
\label{mpjpe}
\end{equation} 
where $N$ is the total number of samples in the dataset , $K$ is the total number of keypoints we are considering for our experiments, $f(x)$ is the predicted 3D pose coordinate by the model and $y$ is our triangulated 3D coordinate used as target data \cite{20202DT3}.


The model operates on keypoints detected in a single time step, \ie $z$ is predicted from the tuple $(x,y)$.
We furthermore added the decaying momentum to the batch normalization layers to make the model more comparable to the \textit{Temporal Convolutional Model}. The momentum parameter decides how much of the statistics of the input variables for a layer is used to normalize the input distribution between the layers. The model was trained using the Adam Optimizer and uses the Kaiming Initializers for initialization of the weights.

\subsection{Temporal Convolutional Model}
\begin{figure}[ht!] 
\centering
\includegraphics[width=0.95\linewidth]{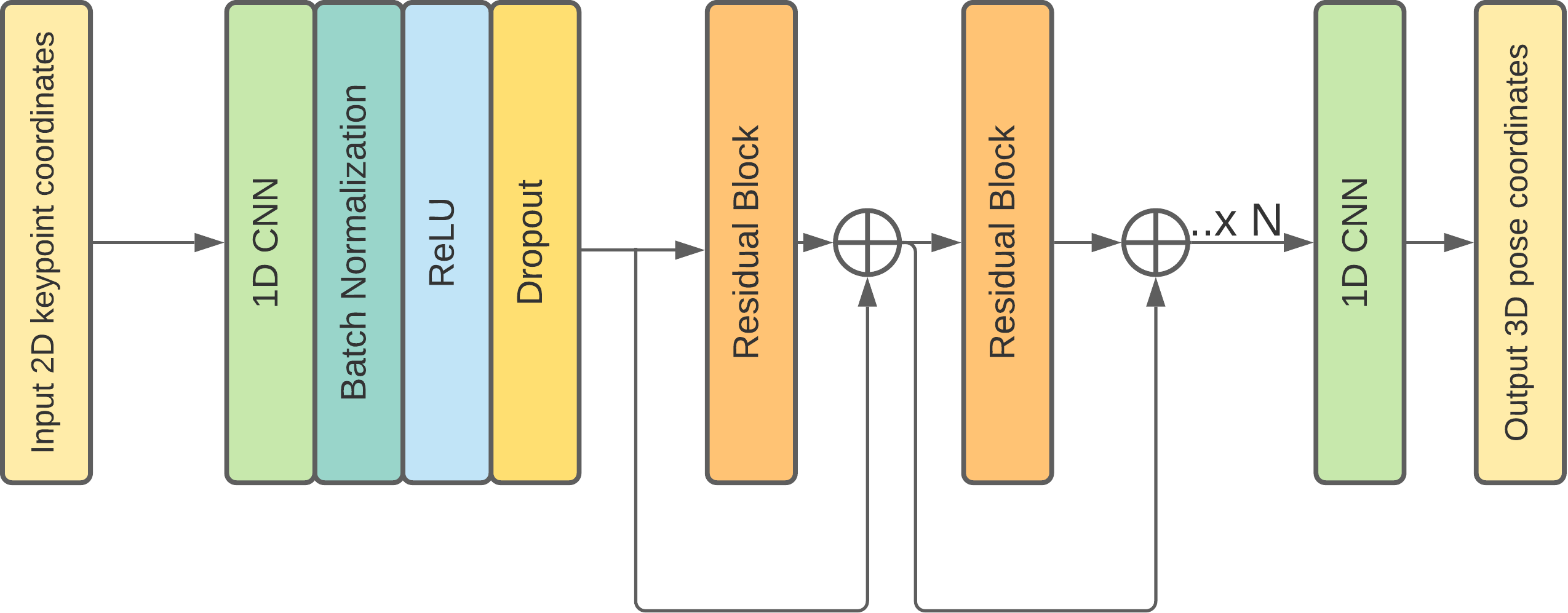}
\caption{Model Architecture of the \textit{Temporal Convolutional model}.}
\label{fig5}
\end{figure}

The \textit{Temporal Convolutional model}\cite{videopose} architecture consists of 1-dimensional convolutional layer followed by a batch normalisation layer, ReLU and dropout (Figure \ref{fig5}). This arrangement is followed by $N$ number of residual blocks where each of consists of a combination of 1-dimensional convolution layer, a batch norm layer, ReLU and dropout followed repeatedly by the same combination again. Finally there exists another 1-dimensional convolutional layer before the output layer. 

As proposed in the original publication \cite{videopose}, the model exists in two variants. One variant of this model uses strided convolutions for training, as it is takes better care of unused intermediate results in the hidden layers.  The other variant uses dilated convolutions which uses all the intermediate results for the prediction of the 3D pose output. In our case, we carried out experiments using strided convolutions both for training and prediction of 3D pose coordinates.
Furthermore, we used symmetric convolutions only in the model, as causal convolutions are rather practical for real-time inference scenarios\cite{videopose}.


For prediction of 3D poses at a particular time step $t$ we consider a temporal window of size $T$ . This temporal window contains input coordinates from timestep $(t-T)$ to timestep $(t+T)$.
Hence as input we consider the past timestep $(t-T)$ together with future timestep $(t+T)$ input coordinates along with the present timestep $t$ coordinates. Hence, the network predicts the current timestep $t$ as output.
The 1-dimensional input kernel convolves vertically along the axis of data samples. The individual keypoints or features for each instance act as different channels for the convolution.
The model was trained using the Adam optimizer and the loss was the MPJPE loss \eqref{mpjpe}.

\section{Results}

In this section we compare the performance of models, the \textit{Linear model} and the \textit{Temporal Convolutional Model}, on the described dataset.
For comparison, we consider three scenarios: the first scenario predicts the $z$ coordinate given the tuple $(x,y)$. The second scenario predicts the $y$ coordinate given the tuple $(x,z)$, resembling the depth inference from the side view of the rodent. In the third scenario also predicts the $y$ coordinate from $(x,z)$ tuples, however the perspective is rotated by $45 \degree$ along the $x$-axis, resembling a diagonal top-down view on the animal as it could be obtained under realistic experimental conditions.

Before using the data we shuffled the data in order to improve the generalization over the test set.
For the \textit{Temporal Convolution model}, we initially divided the data into chunks of the temporal window size and re-arranged them in a random order.
For the \textit{Linear model}, we initially shuffled the data at every time step.
Each model evaluation was repeated for 10 times, where each run was started with random initialization weights and a new random test/train split ($20\%/80\%$).
The evaluation criteria for each model and setting was the final test error that was obtained after 150 training epochs.
Both models were evaluated at the same input data for the MPJPE loss. In both models, the initial learning rate was 0.001 and the decay rate of per epoch was 0.95. 

\begin{figure}[ht!] 
\centering
\includegraphics[width=0.9 \linewidth]{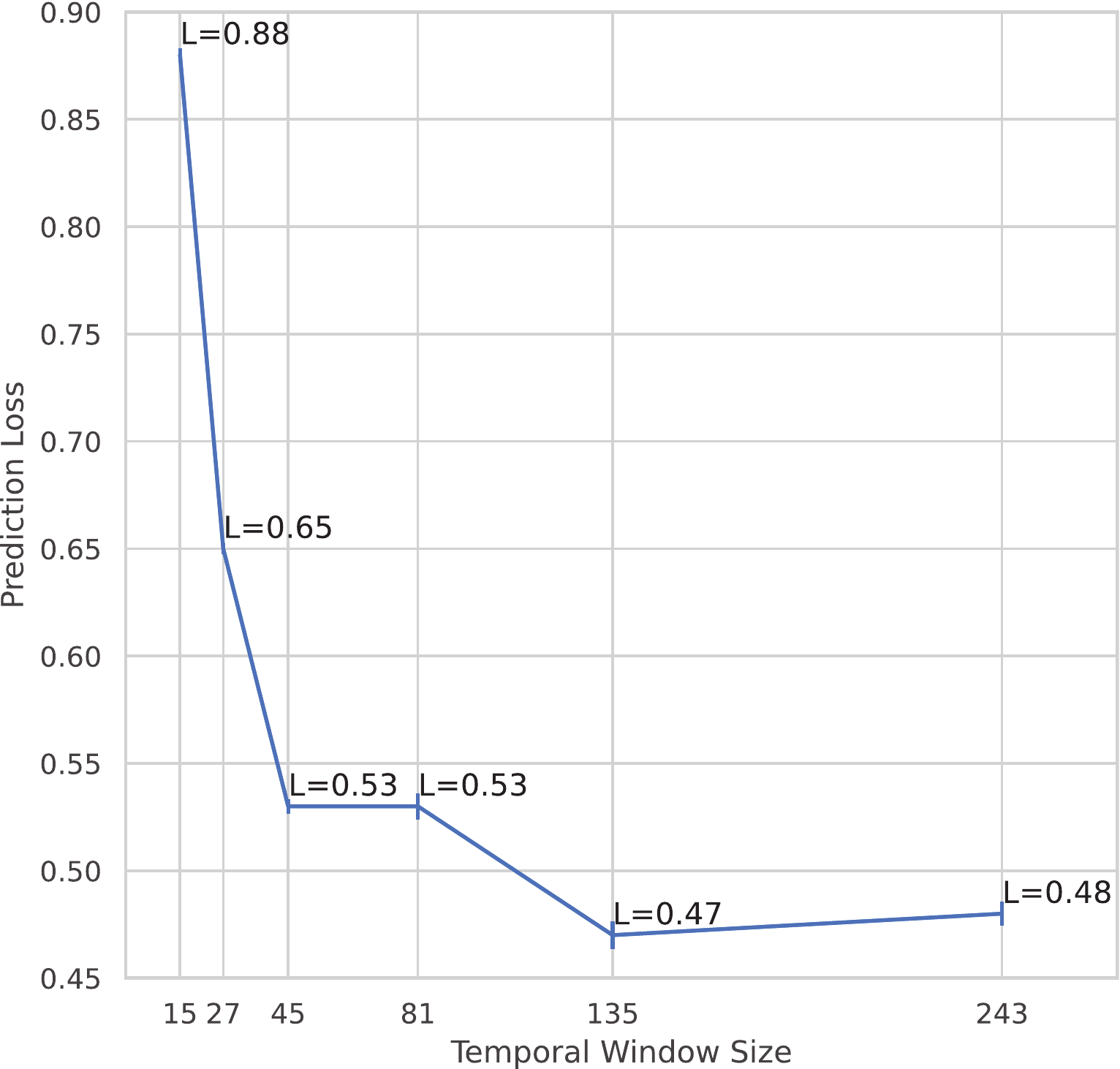}
\caption{Mean test error of the \textit{Temporal Convolutional Model} obtained at increasing temporal window size. The error bars show the loss standard deviation of the 10 shuffled evaluation runs.}
\label{fig6}
\end{figure}

For the \textit{Temporal Convolutional Model} we first determined the optimal temporal window in terms of the test loss.
As shown in Figure \ref{fig6}, the lowest test loss was obtained for window setting 135, which is 2.7 seconds in physical time.
Note that the number of trainable parameters was 6.75 million for the model at temporal window size 15 and 16.8 million for the model at a temporal window size of 243.

\begin{table}[ht]
\centering 
\resizebox{\linewidth}{!}{
\begin{tabular}{p{0.2\linewidth}p{0.2\linewidth}p{0.2\linewidth}}
\hline\hline 
\\Scenario & Linear & Temporal Convolutions \\ [1ex] 
\hline 
\\  $(x,y) \rightarrow z$ & \vtop{\hbox{\strut Train: $4.17 \pm 1.519$}\hbox{\strut Test: $1.47 \pm 0.031$}} & \vtop{\hbox{\strut Train: $0.72 \pm 0.008$ }\hbox{\strut Test: $0.46 \pm 0.006$}} \\  [1ex] 
\hline 
\\ $(x,z) \rightarrow y$  & \vtop{\hbox{\strut Train: $3.40 \pm 0.482$}\hbox{\strut Test: $1.34 \pm 0.064$}} & \vtop{\hbox{\strut Train: $1.01 \pm 0.005$ }\hbox{\strut Test: $0.89 \pm 0.007$}} \\  [1ex] 
\hline 
\\ \vtop{\hbox{\strut$(x,z) \rightarrow y$}\hbox{\strut $45 \degree$ rotation}}  & \vtop{\hbox{\strut Train: $3.17 \pm 0.519$}\hbox{\strut Test: $0.99 \pm 0.037$}} & \vtop{\hbox{\strut Train: $0.61 \pm 0.006$}\hbox{\strut Test: $0.45 \pm 0.003$}} \\  [1ex] 
\hline 
\\  $(z,y) \rightarrow x$  & \vtop{\hbox{\strut Train: $3.18 \pm 0.337$}\hbox{\strut Test: $1.21 \pm 0.025$}} & \vtop{\hbox{\strut Train: $0.69 \pm 0.003$}\hbox{\strut Test: $0.53 \pm 0.004$}} \\  [1ex] 
\hline 
\end{tabular}}
\caption{Performance evaluation of the \textit{Linear model} and \textit{Temporal Convolutional model} for different prediction directions.} 
\label{table:nonlin} 
\end{table}

Finaly, the results for predictions from different perspectives for both models are in Table \ref{table:nonlin}.
In this table, we evaluated the \textit{Temporal Convolutional model} at the optimal window size of 135 timesteps.
We found that under all perspectives, the \textit{Temporal Convolutional model} performed better than the \textit{Linear model}.
Moreoever, we found that both models achieve the lowest test loss at the prediction of $(x,z) \rightarrow y$ (side view along the mouse body), when the perspective is rotated at 45 degrees around the x-axis.
This is followed by the $(x,y) \rightarrow z$ (bottom perspective) for the \textit{Temporal Convolutional model} and the $(x,z) \rightarrow y$ prediction for the \textit{Linear model}.



\section{Conclusion and outlook}
In this work, we present a simple and robust procedure for triangulation of freely moving rodents from multiple cameras that are orthogonal towards a camera positioned underneath the plane of movement. 
Using the triangulated data, we trained and evaluated two types of deep lift posing models that are able to predict the depth coordinate from single camera views, a linear ResNet model and a temporal convolutional ResNet model.
We show that the \textit{Temporal Convolutional model} attained a lower test loss at all viewing angles then the \textit{Linear model}.

During the evaluation of our dataset, we also found that some viewing directions could be more effectively predicted than others.
For example, the diagonal top-down view yielded a lower test-loss than the orthogonal side view. 
We believe that due to the body symmetry some body parts (paws, ears) are hard to distinguish by the network in this perspective.
Interestingly, the $z$ height of body parts given the $(x,y)$ coordinates from the bottom view can be rather efficiently estimated by both models, in comparison to the other directions.

In regards to previous literature, Wiltschko. al. \cite{autocorel} state that the auto correlation of mouse pose dynamics decrease after a period of approximately 500 ms. Interestingly, the optimal temporal window size determined for the \textit{Temporal Convolutional model} was significantly larger (3.7 s).
This suggests that additional behavioral information can be extracted at longer time spans, adding onto the discussion on the relevance of multi-scale dependencies in behavioral models \cite{datta_computational_2019}.

For future work, we aim to implement semi-supervised learning as suggested in \cite{videopose} and \cite{sun_view-invariant_2020} to establish view-invariant pose-lifting for freely moving rodents.
Moreover, we need to evaluate the model under different scenarios, such es occlusions by objects or other animals.
Finally, we hope that this work paves the way to robust 3d pose estimation in complex laboratory environments, allowing for behavior quantification based on 3D pose data for a variety of experimental designs. 

{\small
\bibliographystyle{ieee_fullname}

}

\end{document}